\documentclass[11pt, a4paper]{article}

\usepackage[utf8]{inputenc}
\usepackage{geometry}
\usepackage{graphicx}
\usepackage[colorlinks=true,linkcolor=blue,urlcolor=blue,citecolor=blue]{hyperref}
\usepackage{caption}
\usepackage{float}
\usepackage{url}
\usepackage{booktabs}

\title{\textbf{Fine-Tuning a 7B Advisor on Free-Tier GPUs: An Adapter-Handoff Recipe and a Synthetic-Data Reliability Caution}}

\author{
    \textbf{Md Millat Hosen} \\
    Department of Computer Science and Engineering \\
    Sharda University, Greater Noida, India \\
    \href{mailto:millat6575@gmail.com}{millat6575@gmail.com}
}

\date{}

\begin{document}

\maketitle

\begin{abstract}
Fine-tuning a 7B-parameter language model for a specialized advising task is attractive for resource-constrained settings, but the multi-epoch training runs it requires routinely exceed the wall-clock session limits of the free-tier GPUs (e.g., Kaggle, Colab) available to such users. We report two things. First, a \emph{practical recipe}: a three-epoch QLoRA fine-tune of Mistral-7B-Instruct-v0.3 (4-bit NF4, LoRA rank 16, $\alpha=32$, via Unsloth) completed across two different free-tier 16~GB GPUs (Tesla P100 then Tesla T4) by checkpointing \emph{only} the small LoRA adapter (41.9M parameters) and resuming on the second machine. We document that adapter-only handoff is sufficient---full optimizer and scheduler state need not be transferred---so the binding constraint is per-step VRAM and per-session wall-clock, not the aggregate compute of a single machine. The two-phase run kept peak memory within budget on both GPUs (15.888~GB / 14.741~GB) and reduced training loss from 1.01 to 0.34. Second, and more importantly, an \emph{honest evaluation} of the resulting model that returns a cautionary result. On a blind held-out comparison against the un-fine-tuned base model, the fine-tuned model scored \emph{higher} on reference-based similarity to the (synthetic) training distribution (BERTScore F1 $+0.063$ against synthetic references---a fidelity, not quality, signal) but \emph{lower} on advising quality: a blind LLM-as-judge preferred the base model on 46\% of prompts versus 18\% for the fine-tuned model, and a source-verified factuality audit found four confident factual errors from the fine-tuned model on policy-sensitive topics (healthcare, admissions, and scholarships) against zero for the base model on the same prompts. Fine-tuning on unverified synthetic advising data raised fidelity to the training distribution while degrading factual reliability. Auditing the training data directly with the same source-verified method, we find that this is not a fine-tuning artifact: each of the four audited model errors is already present in the Gemini-generated training answers, and a random-sample audit finds a verifiable factual error in a sizable fraction of training responses (point estimates 28--40\%; single-judge, n=40). The training data is therefore sufficient to account for the observed errors, which we attribute to the synthetic-data pipeline rather than to the adapter-handoff method. We release the dataset, the adapter, the cross-GPU training notebooks, and the full evaluation harness (automatic metrics, the LLM-as-judge scripts, and both the model and dataset factuality catalogs) so that every result can be reproduced from a single 16~GB GPU.
\end{abstract}

\section*{Changes from Earlier Versions (v4)}
\small
This is a substantially revised version of a paper previously circulated under the
title ``A LoRA-Based Approach to Fine-Tuning LLMs for Educational Guidance in
Resource-Constrained Settings'' (arXiv:2504.15610v1--v3). Re-examination of our own
training logs, code, and statistics led us to withdraw two claims made in the
earlier versions and to reframe the contribution honestly:

\begin{itemize}
  \item \textbf{Withdrawn: the ``quantization-boundary'' finding.} Earlier versions
  attributed a loss reduction to a quantization-related effect at a training
  boundary. Our own loss curve is still declining at the end of epoch one, 4-bit
  NF4 quantization is held constant across the claimed boundary, and the additional
  reduction is fully explained by further epochs and a larger effective batch. We
  retract this claim; Section~4.1 now reports the trajectory as ordinary
  three-epoch training behavior.
  \item \textbf{Corrected: cross-GPU state transfer.} Earlier versions implied that
  optimizer and scheduler state were carried across the GPU handoff. The released
  code transfers \emph{only} the LoRA adapter; the optimizer is re-initialized on
  the second GPU. Section~3.3 now states this correctly, and the adapter-only
  handoff is reframed as the operational contribution.
  \item \textbf{Added: an honest reliability evaluation and a training-data audit.}
  A blinded held-out comparison against the un-fine-tuned base model, a blind
  LLM-as-judge, and a source-verified factuality audit show that fine-tuning on
  unverified synthetic data raised fidelity to the synthetic distribution while
  degrading factual reliability. A direct audit of the training data
  (Section~4.4) isolates the cause to the synthetic-data pipeline rather than the
  fine-tuning method.
  \item \textbf{Scope and venue.} The paper is repositioned from an educational-
  guidance contribution to a reproducibility / efficient-ML systems contribution
  with a synthetic-data reliability caution.
\end{itemize}
\normalsize

\section{Introduction}
The demand for accurate, accessible guidance on international study pathways continues to grow, while traditional advising struggles to keep pace with frequently changing admission requirements, scholarship programs, and visa regulations. Large language models (LLMs) are an appealing way to scale such guidance, but their compute and data requirements make them difficult to adopt in the very settings that most need low-cost advising---small college advising offices, public-sector education portals, and institutions outside well-funded research environments. Off-the-shelf general-purpose LLMs also exhibit domain misalignment and can produce fluent but incorrect answers on specialized, policy-sensitive topics.

A natural response is to fine-tune a moderately sized open model on domain data using parameter-efficient methods. In principle, Low-Rank Adaptation (LoRA) with 4-bit quantization (QLoRA) makes this feasible on a single 16~GB consumer or free-tier cloud GPU. In practice, two obstacles confront a resource-constrained practitioner, and this paper is organized around confronting both honestly.

The first obstacle is operational. A multi-epoch fine-tune of a 7B model takes several hours, and the free-tier GPU sessions most accessible to low-resource users (Kaggle, Colab) impose hard wall-clock and weekly-quota limits that a single multi-epoch run can exceed. The binding constraint is therefore not the total compute---which is modest---but the inability to hold one machine for long enough. We show that this can be sidestepped with a simple \emph{cross-GPU adapter-handoff recipe}: train for as long as a session allows, checkpoint only the LoRA adapter, and resume on whatever 16~GB GPU is next available, even a different architecture. We document that transferring only the adapter (not the optimizer or scheduler state) is sufficient to continue training productively.

The second obstacle is evidential, and it is the more important contribution of this paper. Having trained such a model, how does one know it is actually \emph{better} for advising? We constructed an honest, blinded held-out evaluation against the un-fine-tuned base model and found a result we did not expect and do not wish to obscure: fine-tuning on our synthetic advising data made the model score \emph{higher} on reference-based similarity to that synthetic distribution, yet \emph{lower} on independent measures of advising quality and factual accuracy. The fine-tuned model became more willing to state confident, specific---and sometimes wrong---claims on exactly the policy-sensitive topics where errors carry real cost. We report this trade-off as our central finding because it is a transferable caution for the large number of practitioners now fine-tuning small models on synthetically generated data.

We make three contributions:

\begin{enumerate}
    \item \textbf{A free-tier feasibility recipe.} We document a cross-GPU adapter-handoff protocol that completes a three-epoch QLoRA fine-tune of a 7B model on free-tier 16~GB GPUs by spanning the work across sessions and GPU types, keeping peak VRAM within budget on each (15.888~GB on a P100, 14.741~GB on a T4).
    \item \textbf{Documentation that adapter-only handoff suffices across a GPU-architecture boundary.} Continuing training across machines required transferring only the small LoRA adapter (41.9M parameters, on the order of 100~MB on disk); the optimizer and scheduler were re-initialized on the second GPU. That only the adapter carries the learned state is a consequence of the PEFT design rather than a new finding; our contribution is to document that this standard adapter-save workflow is sufficient across a \emph{session and heterogeneous-GPU} boundary at the modest, recoverable cost of an optimizer warm-up, which is the practically relevant case for free-tier users. We do not claim this as a methodological advance.
    \item \textbf{A cautionary reliability finding, with the cause isolated to the data, and the evaluation harness behind it.} On a blind 50-prompt held-out comparison, fine-tuning on unverified synthetic data degraded factual reliability (the base model was preferred 46\% vs.\ 18\% by a blind LLM-as-judge; a source-verified audit found four LoRA factual errors vs.\ zero for the base on policy-sensitive prompts) even as it raised BERTScore. Critically, applying the same audit to the \emph{training data} shows that each of these errors is already present in the synthetic corpus and that a sizable fraction of sampled training answers (point estimates 28--40\%) contain a verifiable factual error---indicating that the data is sufficient to account for the failure and attributing it to the data pipeline rather than the training method. We release the dataset, adapter, training notebooks, evaluation scripts, the LLM-as-judge tooling, and both the model and dataset factuality catalogs.
\end{enumerate}

The remainder of the paper is organized as follows. Section~2 reviews parameter-efficient fine-tuning, quantization, multi-session training, and the reliability risks of synthetic training data. Section~3 describes the dataset, its quality pipeline, the model and hyperparameters, and the two-phase cross-GPU training and evaluation protocols. Section~4 reports the training dynamics, the feasibility and resource results, the held-out reliability evaluation, and---closing the causal loop---a direct audit of the training data that localizes the cause of that reliability result. Section~5 discusses implications and limitations, and Section~6 concludes.

\section{Literature Review}

\subsection{Advances in Large Language Models}
Large language models have advanced rapidly across NLP tasks. Brown et al.~\cite{brown2020} demonstrated strong few-shot in-context learning with GPT-3, and Jiang et al.~\cite{jiang2023} showed that the open-weight Mistral-7B family is competitive with larger models on instruction-following. Despite broad task generalization, these models lack domain specificity and can produce fluent but incorrect output in settings that demand precise, current, contextualized guidance~\cite{ippolito2020}. Study-abroad advising is a demanding case: guidance on visas, scholarships, and admissions is policy-sensitive and time-sensitive, and a general-purpose model with no domain grounding can produce confidently wrong answers.

\subsection{Parameter-Efficient Adaptation and Quantization}
Full fine-tuning of multi-billion-parameter models is infeasible for most groups. LoRA~\cite{hu2021} inserts low-rank trainable matrices into selected linear projections, training only $\sim$0.1--1\% of parameters while leaving base weights frozen; adapters can later be merged for zero-overhead inference. Dettmers et al.~\cite{dettmers2023} introduced QLoRA, combining 4-bit NF4 quantization of the base model with LoRA adapters and paged optimizers to fine-tune a 7B model on a single 16~GB GPU. GPTQ~\cite{frantar2022} addresses inference-time post-training quantization. The Unsloth framework~\cite{unsloth2024} provides kernel-level optimizations and gradient checkpointing for QLoRA. Our training uses QLoRA (NF4 + LoRA) as implemented by Unsloth; we contribute not a new adaptation method but a deployment recipe (a cross-GPU, multi-session schedule) and an evaluation protocol.

\subsection{Multi-Session and Checkpoint-Based Training}
Resuming interrupted training from checkpoints is standard practice supported by frameworks such as the Hugging Face Transformers Trainer~\cite{wolf2020}, which saves optimizer and scheduler state for exact continuation. The PEFT library~\cite{peft2022} formalizes the adapter-level interface (\texttt{save\_pretrained} and \texttt{load\_adapter}) that our handoff procedure uses. In the parameter-efficient setting, however, the trainable state is dominated by the small adapter, which raises a practical question this paper addresses empirically: how much state must actually be carried across a session or machine boundary to continue training productively? We find that, for LoRA, the adapter weights alone suffice; re-initializing the optimizer incurs only a short, recoverable warm-up. This is useful in free-tier environments where session limits, not compute, are the binding constraint.

\subsection{Reliability Risks of Synthetic Training Data}
Generating instruction-tuning data with a strong LLM is now common practice, popularized by approaches such as Self-Instruct~\cite{wang2022}. However, synthetic data can encode the generator's own errors, overconfidence, and stylistic regularities, and training on model-generated content carries documented risks: recursive training on generated data can cause distributional ``model collapse''~\cite{shumailov2024}, and self-consuming training loops degrade quality and diversity over generations~\cite{alemohammad2023}. More directly relevant to a single-round fine-tune, Gudibande et al.~\cite{gudibande2023} show that imitating a stronger model's outputs yields responses that mimic style but not factuality, and Gekhman et al.~\cite{gekhman2024} show that fine-tuning on knowledge a model did not consolidate during pre-training increases its tendency to hallucinate---precisely the hazard when a comparatively weak generation-era model supplies claims outside its reliable knowledge boundary. Our setting is a single, non-recursive fine-tune rather than a feedback loop, but the same underlying hazard applies---a model fine-tuned to match synthetic data may improve on metrics that reward similarity to the generator's distribution while not improving, or even degrading, ground-truth correctness. The confident assertion of specific false facts is itself a well-documented LLM failure mode~\cite{ji2023}. Two further considerations shape our evaluation design. First, reference-based metrics such as BERTScore~\cite{zhang2020}, computed against \emph{synthetic} references, measure fidelity to the generator's distribution rather than factual quality, so evaluation relying on them alone can mislead. Second, LLM-as-judge evaluation~\cite{zheng2023} carries its own biases---most notably self-preference, where a judge favors outputs from its own model family~\cite{panickssery2024}---which we mitigate by using a judge from a different family than the data generator. Accordingly, our evaluation separates these axes: reference-based similarity, a blind LLM-as-judge comparison, and a source-verified factuality audit whose verdicts do not depend on any model's judgment. Our contribution is not to discover the synthetic-data reliability risk but to provide a concrete, source-verified, and causally-audited domain instance of it.

\subsection{AI Applications in Academic Advising}
Rule-based and retrieval-augmented advising systems have been deployed to answer pre-defined academic questions; Cha et al.~\cite{cha2024} review how AI course-recommender systems affect student decision-making, and Kaur et al.~\cite{kaur2022} discuss the limits of rule-based advising. Operational use of LLMs in education is further constrained by energy, model size, and latency~\cite{zhao2023}. LLM-based advising is a natural next step, but the policy-sensitive nature of the domain makes factual reliability---not merely fluency or structure---the property that matters, which motivates our evaluation design.

\section{Methodology}

\subsection{Dataset}
The dataset consists of \textbf{2{,}676 multi-turn student--advisor conversations} (\textbf{2{,}274 training / 402 held-out test}, an 85\%/15\% split), with 6{,}941 user turns and 6{,}941 assistant turns, an average of $5.19 \pm 0.98$ turns per conversation (range 4--6). Every assistant response is expected to follow a markdown structure with \texttt{\#\#} headings and an ``Action Steps'' section. All conversations were generated with the Gemini~1.0~Pro API; no real student--advisor interactions are included. This is a first-class limitation (Section~5.1) and, as Sections~4.3--4.4 show, it is also the subject of our central measurement.

\subsubsection{Topic Distribution}
The dataset spans eight advising topics, balanced by a topic-aware sampling policy. The distribution measured on a 200-sample annotated subset (semantic-similarity classifier, \texttt{sentence-transformers/all-MiniLM-L6-v2}) is shown in Table~\ref{tab:topic-dist}.

\begin{table}[H]
\centering
\caption{Topic distribution over a 200-sample annotated subset.}
\label{tab:topic-dist}
\begin{tabular}{lrr}
\hline
\textbf{Topic} & \textbf{Count} & \textbf{Percent} \\
\hline
other / general advising        & 64 & 32.0\% \\
university / program selection  & 28 & 14.0\% \\
accommodation / living costs    & 25 & 12.5\% \\
student life / cultural adapt.  & 22 & 11.0\% \\
visa / immigration prep.        & 21 & 10.5\% \\
documents / SOP / CV / recs     & 16 &  8.0\% \\
scholarships / funding          & 15 &  7.5\% \\
admissions / application reqs.  &  9 &  4.5\% \\
\hline
\textbf{Total annotated}        & \textbf{200} & \textbf{100.0\%} \\
\hline
\end{tabular}
\end{table}

\subsubsection{Prompt Templates}
Conversations were produced via the Gemini~1.0~Pro API (\texttt{temperature=0.7}, \texttt{top\_p=0.95}, \texttt{top\_k=64}). For each topic, question templates were filled with random parameters drawn from topic-specific value lists. Two prompts drive each conversation:

\begin{itemize}
    \item \textbf{Initial prompt.} ``\textit{You are an experienced study-abroad advisor. A student asks: \texttt{<question>}. Provide a structured response in markdown with section headings and a final `Action Steps' section.}'' This generates the first assistant turn.
    \item \textbf{Follow-up prompt.} ``\textit{You are the same advisor. The student asked: \texttt{<question>}. Your previous answer summarized: \texttt{<context>}. The student follows up: \texttt{<follow-up>}. Continue the conversation in markdown, building on the prior context.}'' This produces 1--4 follow-up turns (yielding the 4--6 turn distribution).
\end{itemize}

Random parameter sampling promotes topical diversity: \texttt{country} is drawn from 30+ destinations, \texttt{degree\_level} from \{Bachelor's, Master's, MBA, PhD, MBBS\}, and \texttt{funding\_source} from \{scholarship, loan, self-funded, employer, mixed\}. Each generated conversation is verified for schema and role alternation before acceptance; failed generations are retried up to three times with exponential backoff before being dropped.

\subsubsection{Quality Pipeline}
Five open-source validation scripts run after generation and produce the metrics in Table~\ref{tab:dataset-quality}:

\begin{enumerate}
    \item \textbf{Schema validation} (\texttt{dataset\_verifier.py}). Asserts each record is a \texttt{\{conversations: [\{from, value\}, \ldots]\}} object with strictly alternating \texttt{human}/\texttt{assistant} roles and no empty fields. \emph{Result: 100\% pass on both splits.}
    \item \textbf{Deduplication} (\texttt{clean\_dataset.py}). Removes exact duplicates by hashing and flags near-duplicates via TF--IDF (unigram+bigram, English stop-words removed) cosine similarity at threshold 0.9. \emph{Result: 0 exact duplicates, 0 near-duplicate pairs.}
    \item \textbf{Repetition audit} (\texttt{clean\_dataset.py}). Detects assistant responses repeated across turns within a conversation. \emph{Result: 2 groups (4 turns) out of 6{,}941 assistant turns in the training split; 0 in test.}
    \item \textbf{Topic balance check} (\texttt{analyze\_quality.py}). Reports the eight-topic distribution. \emph{Result: 32\% `other/general advising'; the seven domain-specific topics account for 68\%.}
    \item \textbf{Training-readiness check} (\texttt{evaluate\_for\_training.py}). Re-tokenizes with the Mistral tokenizer, checks length fit, and verifies EOS alignment.
\end{enumerate}

Crucially, these scripts validate \emph{structure}, deduplication, and tokenization---not factual correctness. No stage of the generation pipeline checks the Gemini-generated claims against authoritative sources, and the question templates are filled by an unconstrained parameter cross-product that can yield factually incoherent prompts (e.g., an ``MS in Data Science at Harvard Medical School,'' which does not exist, or a generic ``Bachelor of Medicine'' label applied indiscriminately to universities that do not use it). Section~\ref{sec:dataaudit} measures the consequence of this design directly.

\begin{table}[H]
\centering
\caption{Structural quality metrics for the released dataset.}
\label{tab:dataset-quality}
\begin{tabular}{lrr}
\hline
\textbf{Metric} & \textbf{Train (2{,}274)} & \textbf{Test (402)} \\
\hline
Schema pass rate                     & 100.0\% & 100.0\% \\
Role alternation pass rate           & 100.0\% & 100.0\% \\
Empty field count                    & 0       & 0       \\
Exact duplicate conversations        & 0       & 0       \\
Near-duplicate pairs (TF--IDF$>$0.9) & 0       & 0       \\
Repeated assistant turns             & 2 grp / 4 turns & 0 grp / 0 turns \\
Distinct-1 (unigram diversity)       & 0.0054  & ---     \\
Distinct-2 (bigram diversity)        & 0.1111  & ---     \\
Mean turns per conversation          & $5.19 \pm 0.98$ & $5.19 \pm 0.98$ \\
\hline
\end{tabular}
\end{table}

The held-out test split is \textbf{never used during training}. For the downstream comparison we draw a fixed, topic-stratified sample of 50 prompts from it (Section~\ref{sec:evaluation}).

\subsection{Model and LoRA Configuration}
The base model is \textbf{Mistral-7B-Instruct-v0.3} (7.24~B parameters), loaded through Unsloth's 4-bit NF4 quantization path (\texttt{unsloth/mistral-7b-instruct-v0.3-bnb-4bit}). The quantized model occupies $\sim$5.5~GB of VRAM, leaving headroom within the 16~GB budget for optimizer state, gradients, and activations. LoRA~\cite{hu2021} adapters are injected into all linear projections (\texttt{q,k,v,o,gate,up,down}) of all 32 transformer layers ($7 \times 32 = 224$ adapted modules). The full configuration is given in Table~\ref{tab:lora-config}.

\begin{table}[H]
\centering
\caption{LoRA adapter and fine-tuning hyperparameters. The same adapter/optimizer configuration is used in both phases; only the per-device batch and gradient-accumulation schedule changes (Section~3.3).}
\label{tab:lora-config}
\begin{tabular}{lll}
\hline
\textbf{Parameter} & \textbf{Value} & \textbf{Notes} \\
\hline
\multicolumn{3}{l}{\emph{LoRA adapter}} \\
LoRA rank ($r$)                      & 16     & \texttt{adapter\_config.json} \\
LoRA $\alpha$                        & 32     & $\alpha / r = 2$ \\
LoRA dropout                         & 0      & disabled in this run \\
Target modules                       & 7 per layer & q,k,v,o,gate,up,down \\
Adapted layers                       & 32     & all transformer blocks \\
Trainable parameters                 & 41{,}943{,}040 & 0.60\% of 7.24~B \\
Bias                                 & none   & \texttt{lora\_bias=False} \\
\multicolumn{3}{l}{\emph{Optimization}} \\
Optimizer                            & AdamW 8-bit & bitsandbytes 8-bit states \\
Learning rate                        & $2 \times 10^{-4}$ & per phase \\
LR scheduler                         & linear & \\
Warmup ratio                         & 0.03   & \\
Max gradient norm                    & 0.3    & gradient clipping \\
Weight decay                         & 0.0    & \\
Mixed precision                      & bfloat16 & via Unsloth \\
Gradient checkpointing               & enabled & \\
\multicolumn{3}{l}{\emph{Tokenization}} \\
Max sequence length                  & 2{,}048 & tokens \\
Padding                              & right & \\
EOS token                            & \texttt{</s>} & Mistral default \\
\hline
\end{tabular}
\end{table}

\subsection{Two-Phase Cross-GPU Training and the Adapter-Handoff Procedure}
Training was performed in two phases on the two free-tier GPUs available to us, with the adapter handed off between them. The phases share the configuration in Table~\ref{tab:lora-config}; only the batch schedule and epoch count differ.

\begin{itemize}
    \item \textbf{Phase 1 (Tesla P100-16GB):} per-device batch 2, gradient accumulation 4, effective batch \textbf{8}. One epoch, 284 optimizer steps, 5~h 47~min 25~s. Peak VRAM 15.888~GB.
    \item \textbf{Phase 2 (Tesla T4-16GB):} per-device batch 4, gradient accumulation 8, effective batch \textbf{32}. Two epochs, 142 optimizer steps, 5~h 26~min 18~s. Peak VRAM 14.741~GB.
\end{itemize}

\paragraph{Handoff procedure (adapter-only).} At the end of Phase 1 we save the LoRA adapter (41.9M parameters, on the order of 100~MB on disk) and push it to a model hub. Phase 2, on a different machine and GPU architecture, re-creates the PEFT model and loads the Phase-1 adapter weights (\texttt{load\_adapter}); it then starts a \emph{fresh} training run. We did \emph{not} transfer the AdamW optimizer state or the LR-scheduler state across the boundary, and Phase 2 does not resume a Hugging Face \texttt{Trainer} checkpoint---only the adapter weights cross the boundary. This is the canonical behavior recorded in the released Phase-2 notebook. The practical consequence is that the per-machine state a practitioner must move between free-tier sessions is just the small adapter, not the full optimizer state; the cost is a brief optimizer warm-up at the start of Phase 2 (Section~4.1). We do not claim the two-phase schedule is a controlled experiment: the GPU, the effective batch size, the optimizer state, and the epoch count all change between phases, so it is a deployment recipe, not an ablation.

\subsection{Training Infrastructure}
\begin{itemize}
    \item \textbf{Phase 1:} NVIDIA Tesla P100 (16~GB VRAM, 3{,}584 CUDA cores, 732~GB/s bandwidth).
    \item \textbf{Phase 2:} NVIDIA Tesla T4 (16~GB VRAM, 2{,}560 CUDA cores, 320~GB/s bandwidth).
    \item \textbf{Software:} PyTorch 2.5.1+cu121, CUDA Toolkit 12.1, Triton 3.1.0, Transformers 4.51.0 (Phase~1) / 4.50.3 (Phase~2), Unsloth 2025.3.19, Weights \& Biases for monitoring. The released \texttt{adapter\_config.json} is the canonical record of the configuration.
\end{itemize}

\subsection{Evaluation Protocol}
\label{sec:evaluation}
We evaluate along two axes: training-time loss dynamics, and downstream generation quality on the held-out split. The downstream protocol compares the LoRA model head-to-head against the un-fine-tuned Mistral-7B-Instruct-v0.3 baseline on identical prompts and generation settings, so the effect of fine-tuning is isolated.

\subsubsection{Held-out Test Set and Generation Settings}
We hold out 402 conversations (15\%) and sub-sample 50 prompts using a fixed seed (\texttt{random.seed(42)}), covering the eight topics in roughly the test-split proportions. Both models are loaded in 4-bit NF4 and decode deterministically (\texttt{do\_sample=False}, \texttt{temperature=0.0}, \texttt{top\_p=1.0}). Both are prompted with the \emph{same} Mistral-Instruct chat template (the base model's native instruction format), so any difference reflects the adapter weights rather than a prompting asymmetry. We evaluate at a 256-token cap and, because 96\% of responses hit that cap, re-run the entire comparison at a 512-token cap (same prompts, seed, models, and chat template); at 512 tokens 24\% of base and 20\% of LoRA responses still truncate, so the reference-based metrics should be read with that caveat.

\subsubsection{Three Complementary Evaluation Signals}
Because reference answers are themselves Gemini-generated, reference-based metrics measure fidelity to the synthetic distribution rather than ground-truth quality. We therefore use three signals that probe different axes:

\begin{enumerate}
    \item \textbf{Reference-based metrics} (SacreBLEU, ROUGE-L F1, BERTScore F1 with \texttt{roberta-large}, \texttt{rescale\_with\_baseline=True}), with 1{,}000-resample bootstrap 95\% CIs on the per-item LoRA$-$base deltas. These measure similarity to the synthetic references.
    \item \textbf{Blind LLM-as-judge}~\cite{zheng2023}. Each pair of responses (base and LoRA) is presented to an LLM judge with the source labels hidden; the judge scores each response independently on domain accuracy (0--3) and helpfulness (0--3) using our rubric, and states a preference. To avoid self-preference bias the judge is from a different model family than the Gemini data generator. This measures perceived advising quality. We disclose its limitations in Section~5: it is a single judge model, single pass, on synthetic-split prompts, and is not a substitute for human evaluation.
    \item \textbf{Source-verified factuality audit.} For prompts where the judge's domain-accuracy scores diverged between models, each contested factual claim is checked against an authoritative external source (government, university, or official program pages). These verdicts do not depend on any model's judgment.
\end{enumerate}

We had originally planned a two-annotator human scoring pass with inter-rater agreement; that human scoring was \emph{not} completed for this submission, and we do not report human scores or an agreement statistic. The LLM-as-judge protocol stands in its place as an automatic, reproducible---but explicitly non-human---signal.

\section{Results}

We report training dynamics (Section~4.1), feasibility and resource use (Section~4.2), and---as the central result---the held-out reliability evaluation (Section~4.3).

\subsection{Training Dynamics Across the Handoff}
\label{sec:dynamics}
\begin{figure}[H]
    \centering
    \includegraphics[width=0.8\textwidth]{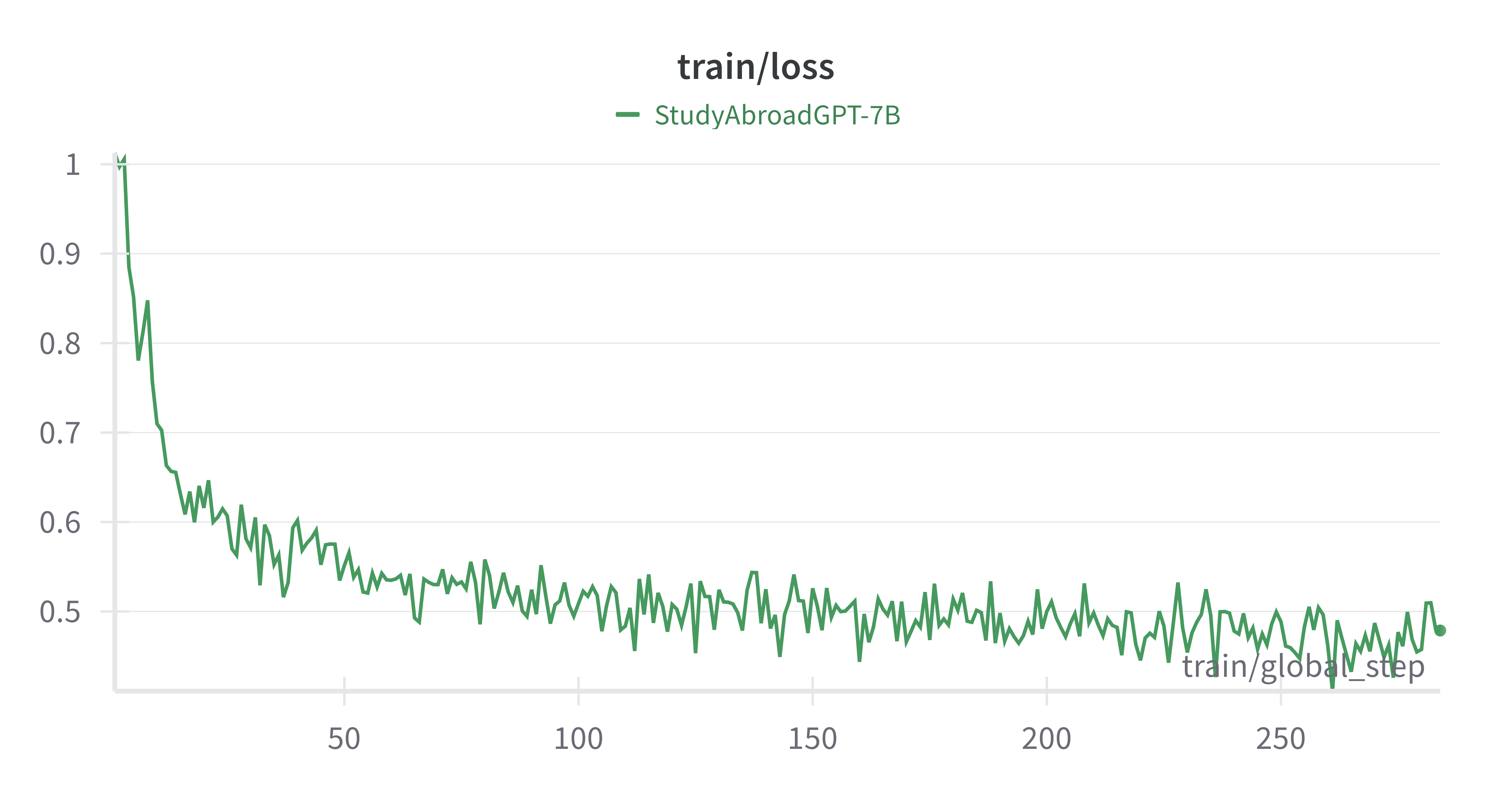}
    \caption{Phase 1 training loss (Tesla P100): $\sim$1.01 to $\sim$0.48 over 284 steps. The curve decelerates but is still declining at end of epoch one.}
    \label{fig:p100_loss}
\end{figure}

\begin{figure}[H]
    \centering
    \includegraphics[width=0.8\textwidth]{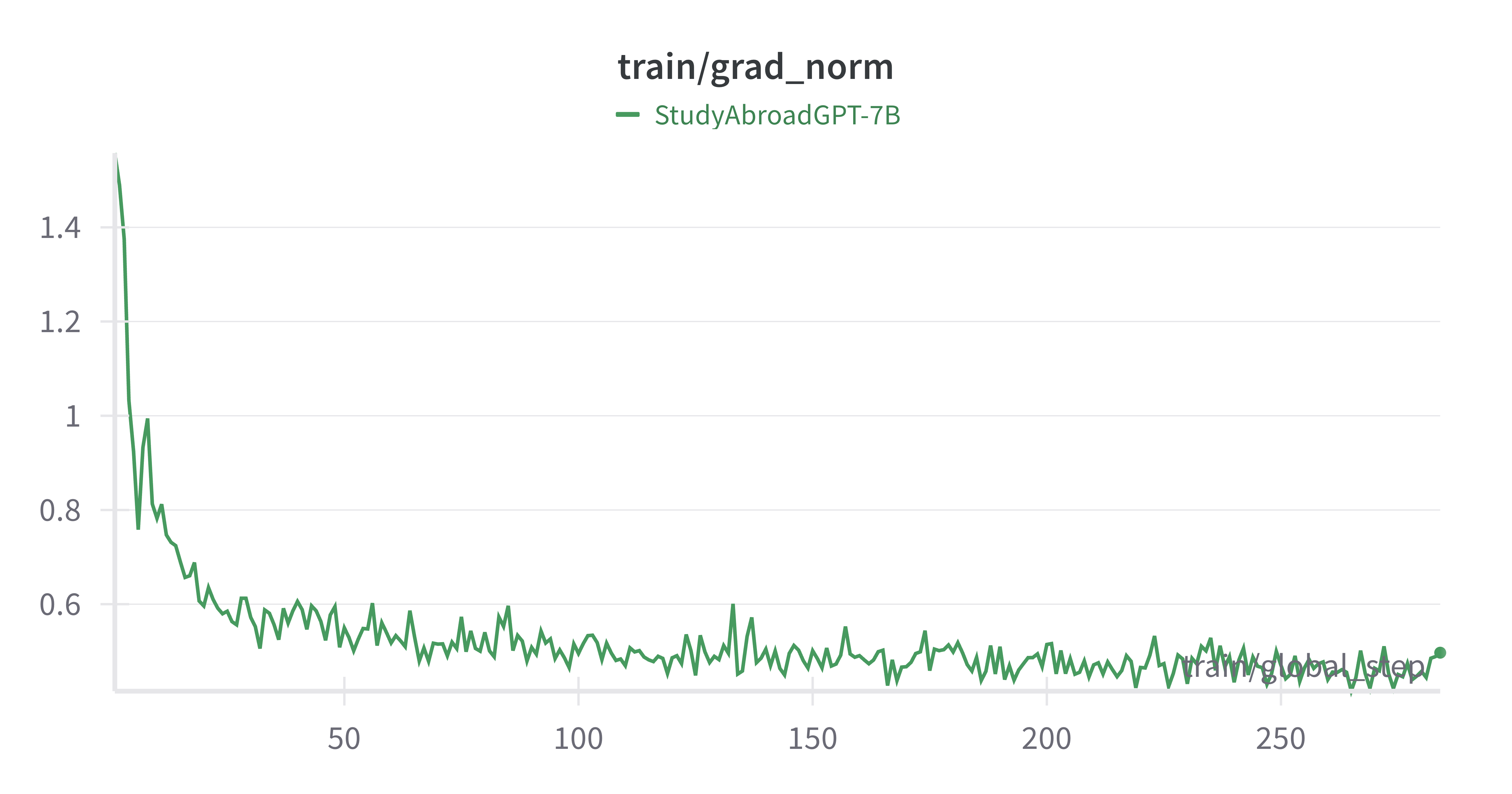}
    \caption{Phase 1 gradient norm (Tesla P100): stabilizes after the first $\sim$50 steps.}
    \label{fig:p100_grad}
\end{figure}

Phase 1 (P100) began at training loss \textbf{1.0125} and ended at \textbf{0.4787} after 284 optimizer steps (one epoch over 2{,}274 conversations), with the lowest single-step value of 0.4109 reached at step 261. The loss curve (Figure~\ref{fig:p100_loss}) decelerates over the epoch but does not flatten: across the final 90 steps the trend is still downward at roughly $-2.7\times10^{-4}$ loss/step. We therefore make no claim that the $\sim$0.48 region is a floor; it is simply where one epoch of training lands. Gradient norms (Figure~\ref{fig:p100_grad}) stabilize after the first $\sim$50 steps.

\begin{figure}[H]
    \centering
    \includegraphics[width=0.8\textwidth]{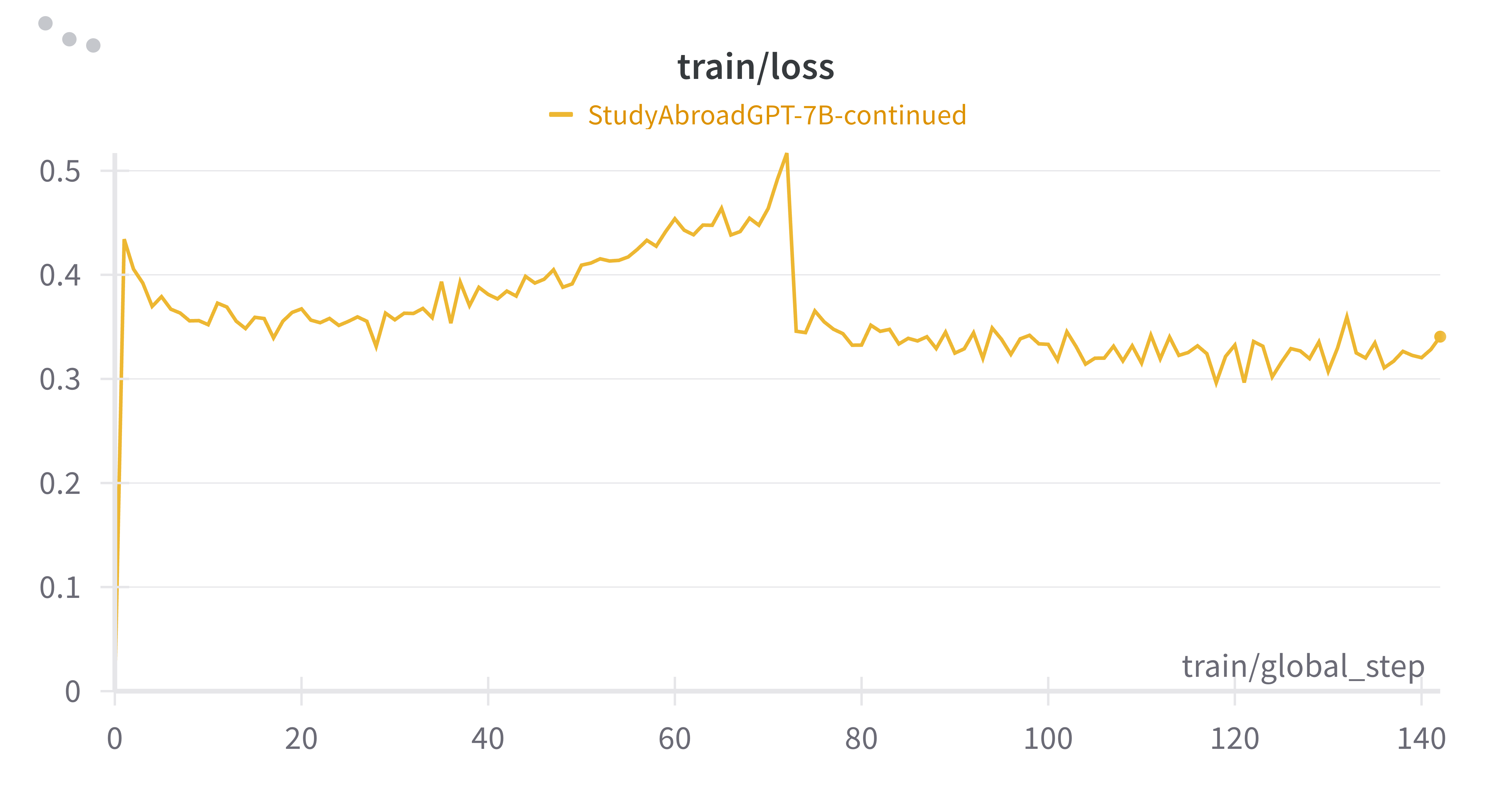}
    \caption{Phase 2 training loss (Tesla T4): continuing from the loaded adapter, $\sim$0.43 to 0.34 over 142 steps.}
    \label{fig:t4_loss}
\end{figure}

\begin{figure}[H]
    \centering
    \includegraphics[width=0.8\textwidth]{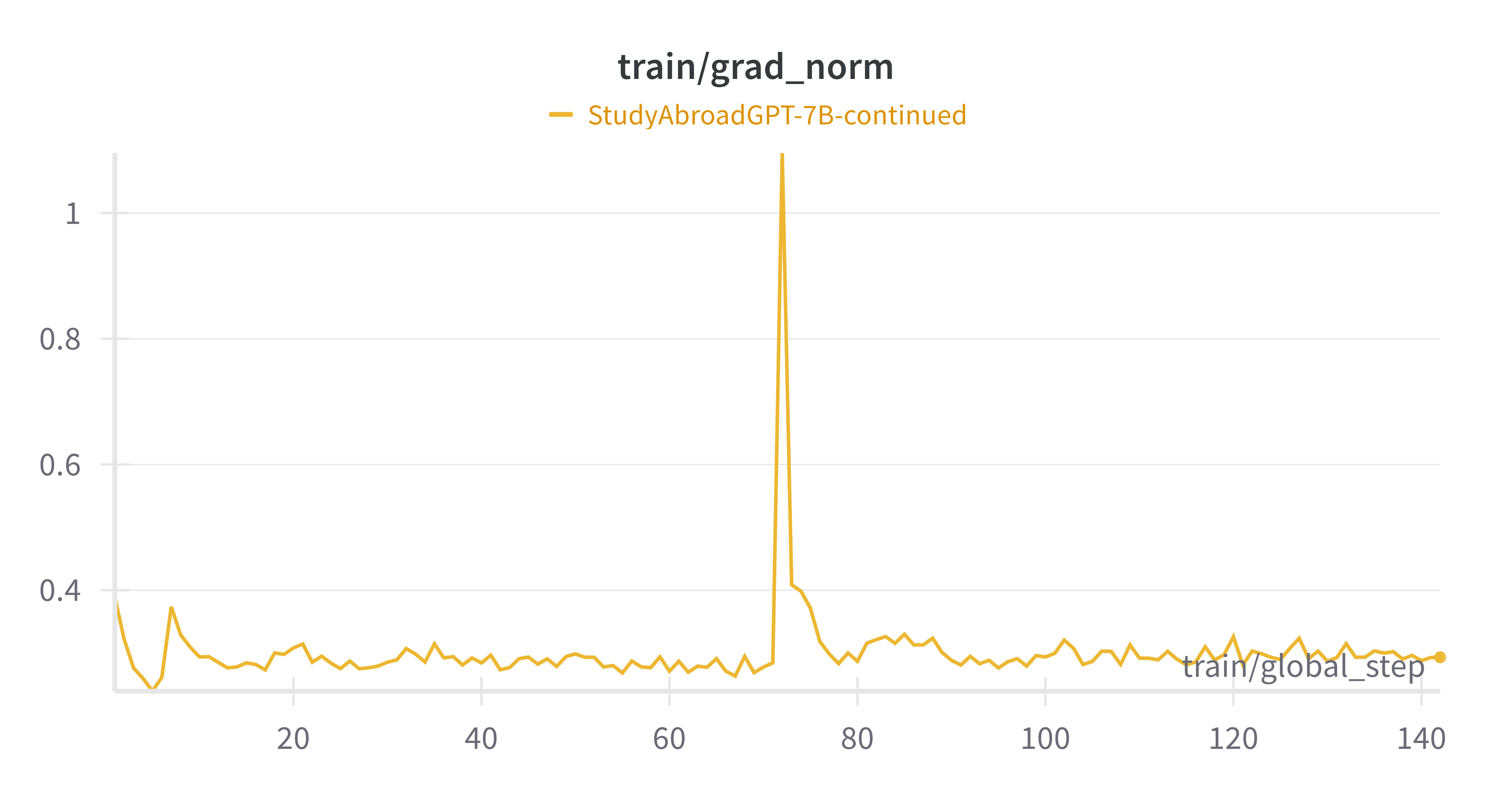}
    \caption{Phase 2 gradient norm (Tesla T4): stable throughout.}
    \label{fig:t4_grad}
\end{figure}

Phase 2 (T4) loaded the Phase-1 adapter and, with a fresh optimizer, ran two further epochs (142 steps), beginning around training loss 0.43 and ending at \textbf{0.3405} (lowest single-step 0.2963). The brief early-Phase-2 plateau is consistent with re-initializing the AdamW state, which must re-estimate its moments before resuming effective descent---the expected, recoverable cost of the adapter-only handoff. The overall two-phase reduction was from 1.01 to 0.34. Gradient norms remained stable throughout Phase 2 (Figures~\ref{fig:t4_loss}, \ref{fig:t4_grad}), so the additional reduction is not driven by instability. We emphasize that this 1.01$\rightarrow$0.34 trajectory is ordinary three-epoch training behavior, with the larger Phase-2 effective batch (32 vs.\ 8) giving a smoother gradient estimate; we do not attribute it to any quantization-specific effect.

\subsection{Feasibility and Resource Use}
\begin{figure}[H]
    \centering
    \includegraphics[width=0.9\textwidth]{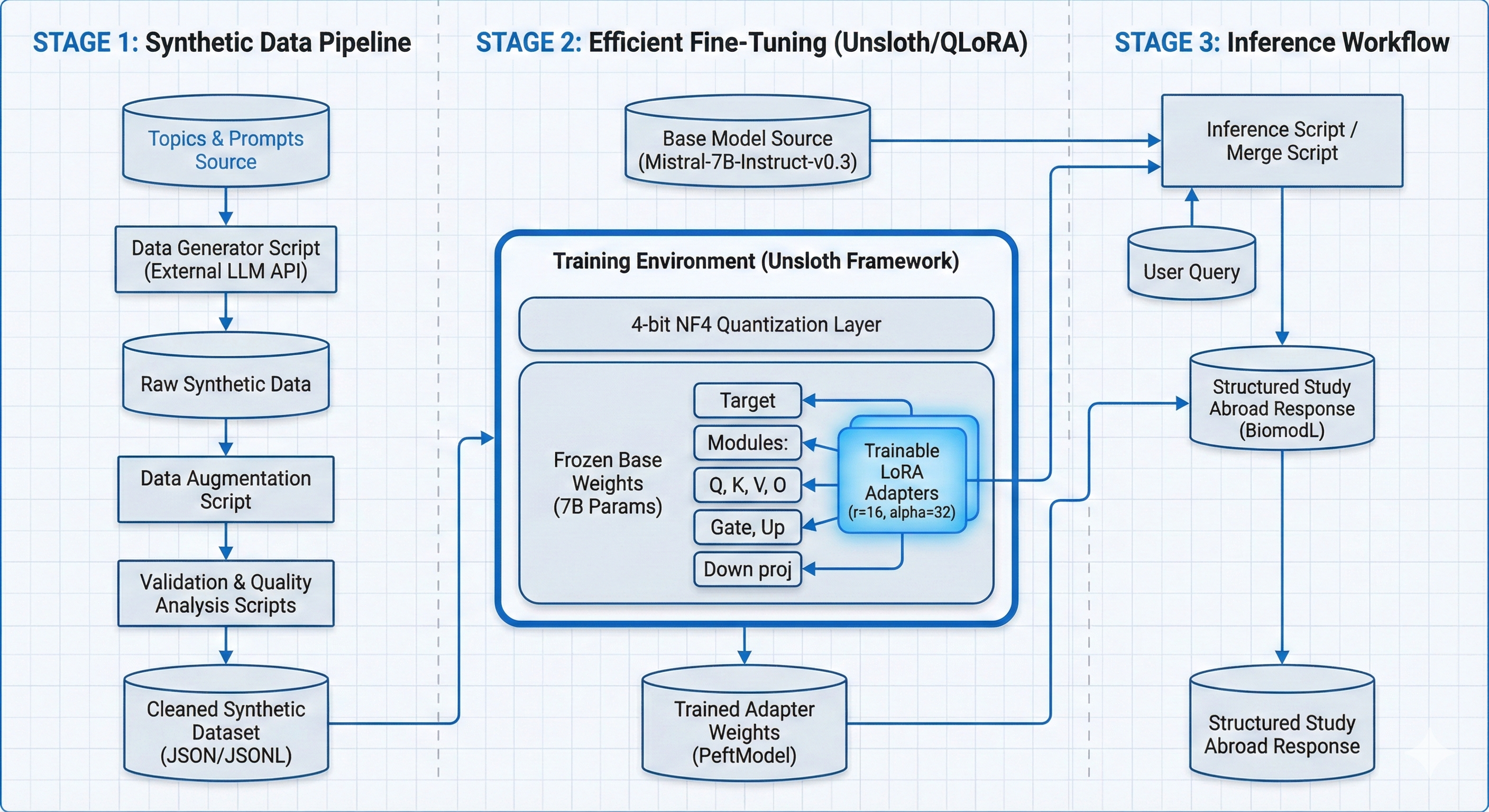}
    \caption{Pipeline architecture. Dataset construction and quality-control scripts (left, Section~3.1) feed the training pipeline (right), which produces a LoRA adapter consumed by the evaluation harness (Section~3.5). Each labeled box corresponds to a released script.}
    \label{fig:architecture}
\end{figure}

The recipe's value is operational. Both phases stayed within the 16~GB budget---\textbf{P100 peak 15.888~GB}, \textbf{T4 peak 14.741~GB}---and each phase fit within a single free-tier session ($\sim$5.5~h). By handing off only the adapter, a three-epoch run that no single free-tier session could complete was finished across two sessions on two different GPU architectures. The binding constraint for fine-tuning a 4-bit-quantized 7B model in these environments is thus per-step VRAM and per-session wall-clock, not the aggregate compute of one machine; a single 16~GB budget can be shared across a heterogeneous free-tier fleet. The architecture of the full pipeline---dataset construction and quality control feeding training, which produces the adapter consumed by evaluation---is shown in Figure~\ref{fig:architecture}.

\subsection{Held-out Reliability Evaluation: a Fidelity--Reliability Trade-off}
\label{sec:reliability}
This is the central result. We present the three signals in order of evidential strength, ending with the reference metrics, because the reference metrics are the most easily misread.

\subsubsection{Source-Verified Factuality Audit}
On the 50-prompt held-out set, the blind LLM-as-judge (Section~4.3.2) assigned different domain-accuracy scores to the two models on 18 prompts; of these, \textbf{16 favored the base model and 2 favored the LoRA model}. We audited the highest-sensitivity divergences against authoritative external sources. On four prompts the LoRA model produced a confident factual error where the base model did not (Table~\ref{tab:factuality}). The two prompts where the LoRA model scored higher were completeness/specificity gains, not corrections of any base factual error.

\begin{table}[H]
\centering
\caption{Source-verified factual errors: LoRA model vs.\ base model on the same prompts. Verdicts are checked against external sources and do not depend on any model's judgment.}
\label{tab:factuality}
\begin{tabular}{p{2.2cm}p{4.6cm}p{4.6cm}}
\hline
\textbf{Topic} & \textbf{LoRA (fine-tuned) --- incorrect} & \textbf{Base --- correct} \\
\hline
Harvard Medical School testing & States HMS ``requires all applicants to submit GRE scores.'' & States HMS requires the MCAT and does not require the GRE. \\
Australian healthcare for intl.\ students & States students are ``eligible for Medicare \ldots after \ldots at least six months.'' & States intl.\ students are not eligible for Medicare and must hold OSHC. \\
``Bachelor of Medicine at Stanford'' & Elaborates admission steps for a program that does not exist. & States Stanford offers no direct undergraduate medicine degree. \\
Scholarships for Brazilian students in Bangladesh & Lists specific named scholarships not established to exist. & States such scholarships are not common; gives honest alternatives. \\
\hline
\end{tabular}
\end{table}

These four cases fall on healthcare, admissions, and scholarship topics---policy-sensitive areas where incorrect advice carries real cost. We present them as illustrative verified cases rather than a statistical sample: with only four errors we make no claim about their distribution across topics. Their common form is the confident assertion of a specific (false) fact, consistent with fine-tuning on synthetic data that is fluent and structurally confident but not fact-checked.

To rule out the most obvious confound---that the base model ``wins'' merely by declining to answer---we counted abstentions and hedged non-answers (e.g., ``I am not sure,'' ``consult the official source'') across the 50 prompts: the base model abstained on 2/50 and the fine-tuned model on 1/50. The base model's advantage is therefore not explained by refusal; both models answer substantively on essentially all prompts, and the difference is in the correctness of the substantive answers.

\subsubsection{Blind LLM-as-Judge}
A blind LLM-as-judge (a single Claude Sonnet-family model, different from the Gemini generator; pointwise absolute scoring per our 0--3 rubric, with source labels hidden) scored all 50 pairs. We use pointwise rather than pairwise scoring to reduce order/position bias, and a cross-family judge to reduce self-preference bias~\cite{panickssery2024}; the exact judge model identifier is recorded in the released harness. We flag two dependencies here and revisit them in Section~5.2: the result rests on a single judge and a single pass (so the bootstrap CIs below capture prompt-level, not judge-level, variance), and the same judge family is later used in the training-data audit (Section~\ref{sec:dataaudit}), so those two results are not statistically independent. Results are in Table~\ref{tab:judge}. The base model was preferred on \textbf{23/50 (46\%)} prompts, the LoRA model on \textbf{9/50 (18\%)}, with 18/50 (36\%) ties. Both per-dimension deltas favor the base model with bootstrap 95\% CIs that exclude zero. The judge's written rationales align with the factuality audit (confident LoRA errors on specific facts).

\begin{table}[H]
\centering
\caption{Blind LLM-as-judge scores on the 50-prompt held-out set (0--3 scales). Deltas are LoRA$-$base; CIs are 1{,}000-resample bootstrap over per-item deltas and capture prompt-level, not judge-level, variance (single judge, single pass; see Section~5.2).}
\label{tab:judge}
\begin{tabular}{lrrrl}
\hline
\textbf{Metric} & \textbf{Base} & \textbf{LoRA} & \textbf{$\Delta$ (LoRA$-$Base)} & \textbf{95\% CI on $\Delta$} \\
\hline
Domain accuracy (0--3) & $2.14 \pm 0.53$ & $1.74 \pm 0.69$ & $-0.40$ & [$-0.62$, $-0.18$] \\
Helpfulness (0--3)     & $2.14 \pm 0.49$ & $1.82 \pm 0.59$ & $-0.32$ & [$-0.52$, $-0.12$] \\
\hline
\end{tabular}
\end{table}

\subsubsection{Reference-Based Metrics (512-token re-run)}
Because 96\% of responses hit the 256-token cap, we re-ran the comparison at 512 tokens (truncation then 24\% base / 20\% LoRA). On reference-based metrics the LoRA model scores \emph{higher} than the base model (Table~\ref{tab:std-metrics}): SacreBLEU $+3.34$, ROUGE-L F1 $+0.019$, BERTScore F1 $+0.063$, the latter two with 95\% CIs excluding zero.

\begin{table}[H]
\centering
\caption{Reference-based metrics on the 512-token re-run. Reference answers are the Gemini-generated test-split responses, so these measure \emph{fidelity to the synthetic distribution}, not ground-truth quality.}
\label{tab:std-metrics}
\begin{tabular}{lrrrl}
\hline
\textbf{Metric} & \textbf{Base} & \textbf{LoRA} & \textbf{$\Delta$ (LoRA$-$Base)} & \textbf{95\% CI on $\Delta$} \\
\hline
SacreBLEU (corpus)         & 5.71   & 9.05   & $+3.34$   & corpus-level (single value) \\
ROUGE-L F1 (per-item mean) & 0.1937 & 0.2125 & $+0.0188$ & [+0.0098, +0.0284] \\
BERTScore F1 (rescaled)    & 0.0981 & 0.1611 & $+0.0631$ & [+0.0473, +0.0785] \\
\hline
\end{tabular}
\end{table}

\subsubsection{Synthesis}
The three signals are consistent once read correctly. Fine-tuning moved the model \emph{toward} the synthetic training distribution (higher BERTScore/ROUGE/SacreBLEU against synthetic references), and \emph{because} that distribution contains fluent but unverified, sometimes-incorrect advice, moving toward it \emph{lowered} factual reliability and judged advising quality (factuality audit and LLM-as-judge both favor the base model). The reference-based gains and the reliability losses are not in tension: they are the same phenomenon measured against two different targets---the synthetic distribution versus ground truth. Reporting BERTScore alone would have inverted the conclusion. The practical lesson is that fine-tuning a small model on unverified synthetic data can trade factual reliability for distribution fidelity, and that distribution-similarity metrics must not be the sole basis for claiming improvement on a policy-sensitive task. These signals establish \emph{that} fine-tuning lowered reliability; Section~\ref{sec:dataaudit} establishes \emph{why}, by auditing the training data directly.

\subsection{Auditing the Training Data: Isolating the Cause}
\label{sec:dataaudit}
The signals above show that fine-tuning lowered reliability, but not whether the fault lies in the training procedure or in the data it was trained on. To separate these, we apply the same source-verified, judgment-independent method to the training corpus itself. The dataset was generated in 2024 with Gemini~1.0~Pro---a comparatively weak generation-era model---via the unconstrained template cross-product described in Section~3.1, with no factuality-checking stage.

\subsubsection{The Model's Errors Are Present in the Training Data}
For each of the four audited model errors (Table~\ref{tab:factuality}), we located training examples on the same topic and checked the Gemini-generated answers against the same authoritative sources. The match is at the level of \emph{error class} rather than the identical prompt string: for instance, the model's confabulation of a nonexistent ``Bachelor of Medicine at Stanford'' corresponds to the same template-generated nonexistent-program elaboration in the training data, there instantiated for other universities (``Bachelor of Medicine at Oxford,'' ``MS in Data Science at Harvard Medical School''). With that reading, all four error classes appear in the training data in the same form (Table~\ref{tab:dataset-causal}). The clearest case is Australian healthcare: a training answer states that international students ``who have been living in Australia for more than 12 months may be eligible'' for Medicare, whereas international students are in general \emph{not} eligible and must hold Overseas Student Health Cover (OSHC) as a visa condition;\footnote{\url{https://www.studyaustralia.gov.au/en/plan-your-move/overseas-student-health-cover-oshc}; \url{https://www.privatehealth.gov.au/health_insurance/overseas/overseas_student_health_cover.htm}} the fine-tuned model reproduces this same false-eligibility claim (it states a different qualifying period, but the same incorrect premise that a residency duration confers Medicare eligibility). The training data likewise treats incoherent, template-generated programs as real and elaborates fabricated requirements, and it invents scholarship programs for corridors where none are established---for example, naming ``Commonwealth Scholarships \ldots including Brazil,'' although Brazil is not a Commonwealth member.\footnote{\url{https://study-uk.britishcouncil.org/scholarships-funding/commonwealth-scholarships}} The model did not invent these failure modes; it inherited them.

\begin{table}[H]
\centering
\caption{Causal match: each audited model error traces to an error already present in the Gemini-generated training data. Verdicts are checked against the same external sources used in Table~\ref{tab:factuality}.}
\label{tab:dataset-causal}
\begin{tabular}{p{2.4cm}p{5.0cm}p{4.4cm}}
\hline
\textbf{Topic} & \textbf{Error present in training data} & \textbf{Ground truth} \\
\hline
Australian healthcare & Training answer: intl.\ students ``eligible \ldots after \ldots 12 months'' for Medicare. & Not eligible; OSHC compulsory under visa condition. \\
Brazil$\rightarrow$Bangladesh scholarships & Invents ``Commonwealth Scholarships \ldots including Brazil.'' & Brazil is not a Commonwealth member. \\
``Bachelor of Medicine at Oxford'' & Elaborates entry requirements for a template-generated program label. & Oxford's degree is the BM~BCh; the label is a template artifact. \\
``MS in Data Science at Harvard Medical School'' & Answers as if the program exists; fabricates requirements. & No such program exists at HMS. \\
\hline
\end{tabular}
\end{table}

\subsubsection{Roughly a Quarter to a Half of the Training Data Contains Verifiable Errors}
To gauge how widespread such errors are, we drew a random sample of 40 training answers (two independent seeds) and scored each for verifiable factual errors using the same blind LLM-as-judge, anchoring every contested claim to an external source. Counting only hard, clearly false or fabricated claims, \textbf{11/40 (27.5\%)} answers contained at least one error (Wilson 95\% CI [16\%, 43\%]); on an inclusive count that also admits medium-confidence errors (e.g., a wrong accreditation body, a fabricated score cutoff), the rate is \textbf{16/40 (40\%)} (CI [26\%, 55\%]). Because this rests on a single judge at a modest sample size, we report it as \emph{indicative}---point estimates of 28--40\%, with the union of the Wilson intervals spanning 16--55\%---rather than as a precise rate. Two caveats apply: the judge's own verification can err, and we did not human-validate its per-answer verdicts, so the point estimate could be biased in either direction; and the same judge family produced the model evaluation (Section~4.3.2), so the two error signals are not independent. The judgment-\emph{independent} causal match (Section~4.4.1) does not depend on this rate. The dominant failure mode is the confident fabrication of specific, citable-sounding facts: invented minimum scores, statistics attributed to real organizations, and program details---the signature of a comparatively weak generation-era model (Gemini~1.0~Pro) prompted with an unvalidated template cross-product and no factuality gate.

\subsubsection{The Data Is Sufficient to Explain the Errors}
Together these results show that the training data is \emph{sufficient} to account for the observed errors: each failure the model exhibits is present in the corpus it learned from, and that corpus carries verifiable errors in a sizable fraction of answers. Distribution-faithful fine-tuning did exactly what it should---it moved the model toward its training distribution (Section~4.3.3)---so inheriting that distribution's errors is the expected consequence, not an artifact of the adapter-handoff procedure. We therefore attribute the reliability drop to the synthetic-data pipeline rather than to the training method. We are deliberately careful about the strength of this claim: a single fine-tuning run cannot establish that the method is sound \emph{in general}, and the causal match shows the data is sufficient to produce the errors, not that the method contributes nothing. The decisive test we have not run---and flag as the key follow-up (Section~5.3)---is to compare base and fine-tuned error rates on prompts whose training neighbors are \emph{verified correct}: if fine-tuning degrades reliability only where the training data is wrong, and not where it is right, the attribution to the data would be established rather than merely supported. What we can claim now is that an identical recipe applied to a factuality-gated corpus would be expected to behave very differently, and that nothing in the evidence implicates the adapter-handoff method itself.

\section{Discussion}

\subsection{Implications}
For practitioners fine-tuning small models on synthetic instruction data---an increasingly common, low-cost strategy---our central result is a caution: similarity to the generator's distribution can rise while ground-truth reliability falls, and the two can be confused if evaluation relies on reference-based metrics computed against synthetic references. On policy-sensitive tasks, evaluation should include at least one judgment-independent factuality signal. Our dataset audit (Section~\ref{sec:dataaudit}) makes the mechanism concrete rather than hypothetical: a weak generation-era model (Gemini~1.0~Pro), an unconstrained template cross-product, and the absence of any factuality gate produced a corpus in which roughly a quarter to a half of answers are factually wrong, and faithful fine-tuning inherited those errors. A corollary for system design is that synthetic advising data should be factuality-gated before training, or paired with retrieval against authoritative sources at inference time, rather than used to teach the model to assert specifics confidently.

We also state the method-versus-data verdict explicitly, because it bears on how this result should be read: the reliability drop is a property of the \emph{data}, not of the adapter-handoff recipe or LoRA fine-tuning. The same recipe over a fact-gated corpus would be expected to improve reliability. The systems contribution (Section~4.2) is therefore not undermined by the reliability finding---the two are independent.

For the operational recipe, the implication is more straightforward: free-tier session limits, not compute, are the binding constraint for multi-epoch 7B QLoRA, and adapter-only handoff is a simple, low-overhead way around them. We stress, however, that the two contributions compose into a hazard if read carelessly: the recipe lowers the cost of producing a domain advisor, and our finding shows that the cheapest data source---unverified synthetic generation---degrades exactly such an advisor. We therefore offer the recipe \emph{with} a precondition: for any advising use, the training corpus must be factuality-gated or paired with retrieval against authoritative sources before fine-tuning. This matters most for the population this work targets. The harm is concrete: a confidently wrong claim that international students are ``eligible for Medicare after six months'' could lead a student to forgo compulsory health cover and breach a visa condition, and elaborated admission steps for a nonexistent program waste the scarce time and money of an applicant who can least afford it. The equity concern is sharp---resource-constrained students, the intended beneficiaries, are also those least able to cross-check advice or absorb the cost of acting on it, so an unreliable advisor inverts the tool's purpose.

\subsection{Limitations}
\begin{enumerate}
    \item \textbf{Training and evaluation data are fully synthetic.} All 2{,}676 conversations were generated by Gemini~1.0~Pro; no real student--advisor interactions are included, and the held-out references are themselves synthetic. This bounds what the evaluation can establish about real-world advising and is the precondition for the reliability finding itself.
    \item \textbf{The LLM-as-judge is a single model, single pass, on synthetic-split prompts.} It is an automatic, reproducible signal, not human evaluation; a different judge family or human raters could differ. We did not complete the originally planned two-annotator human scoring, and we report no inter-rater agreement statistic.
    \item \textbf{The factuality audit is curated, not exhaustive.} It verifies the highest-sensitivity divergent cases against external sources; a full per-claim audit of all 100 responses (50 prompts $\times$ 2 models) remains future work. The reported direction (16/18 divergent prompts favoring the base model) is, however, one-sided.
    \item \textbf{The dataset prevalence rate rests on a single judge at modest sample size.} The training-data error rate (Section~\ref{sec:dataaudit}) is estimated from 40 randomly sampled answers scored by one LLM judge; the resulting Wilson intervals are wide ([16\%, 43\%] hard count), so we report it as indicative (point estimates 28--40\%) rather than a precise figure. The \emph{causal} match (each model error present in the training data) does not depend on this rate. A larger pass with multiple judges and inter-rater agreement is future work.
    \item \textbf{The two LLM-judged signals are not independent.} The same judge family scored both the model comparison (Section~4.3.2) and the training-data audit (Section~\ref{sec:dataaudit}); a shared judge bias would correlate the two results and inflate their apparent coherence. The judgment-independent signals---the source-verified model factuality audit and the source-verified causal match---are the load-bearing evidence for the central claims, and we foreground them accordingly. We additionally note that the factuality-audit set is judge-derived (the divergent prompts are defined by where the judge's scores differed), so the audit corroborates the \emph{direction} of the judge's verdict rather than providing a fully independent check.
    \item \textbf{The reliability finding rests on a single fine-tuning run.} We trained one adapter and evaluated one 50-prompt sample; the reported CIs capture prompt-level variance, not training-run (seed) variance. Replication across seeds would strengthen the result, and our claims about the method are scoped accordingly (Section~4.4.3).
    \item \textbf{Reference-based metrics use synthetic references.} The SacreBLEU/ROUGE/BERTScore values measure fidelity to the synthetic distribution, not ground-truth quality, and should not be read as quality gains.
    \item \textbf{The two-phase schedule is a recipe, not an ablation.} GPU, effective batch size, optimizer state, and epoch count all change between phases, so the run does not isolate the effect of any single factor; we make no causal claim about quantization or batch size.
    \item \textbf{Single language and domain.} The dataset is English-only and study-abroad-specific; transfer is not established.
\end{enumerate}

\subsection{Future Work}
The most consequential next steps are: (i) collecting a small set of redacted, IRB-approved \emph{real} student--advisor interactions to serve as a non-synthetic test set and to support a genuine human evaluation with inter-rater agreement; (ii) confirming the reliability finding with a second, independent judge model, a full systematic factuality audit, and a human spot-check of the dataset-audit judge's verdicts; (ii$'$) the decisive method-vs-data test of Section~4.4.3---comparing base and fine-tuned error rates on prompts whose training neighbors are verified correct versus verified wrong---which would upgrade the data attribution from \emph{supported} to \emph{established}; and (iii) testing whether factuality-gating the synthetic training data, or adding retrieval against authoritative sources, recovers reliability without sacrificing the structural improvements fine-tuning provides. Items (i) and (ii) are not claimed as results here.

\section{Conclusion}
We presented a free-tier recipe for completing a multi-epoch 7B QLoRA fine-tune by checkpointing only the LoRA adapter and handing it off across heterogeneous 16~GB GPUs, keeping peak VRAM within budget on each and reducing training loss from 1.01 to 0.34. We then evaluated the resulting model honestly and found a cautionary trade-off: fine-tuning on unverified synthetic advising data increased fidelity to the training distribution (BERTScore F1 $+0.063$) while degrading factual reliability, with the base model preferred 46\% vs.\ 18\% by a blind LLM-as-judge and producing zero source-verified factual errors against the LoRA model's four on policy-sensitive prompts. Auditing the training data with the same method, we showed that the training data is sufficient to account for this reliability drop: the model's errors are already present in the Gemini-generated corpus, of which a sizable fraction of sampled answers (point estimates 28--40\%) are factually wrong, which attributes the failure to the synthetic-data pipeline rather than the fine-tuning method. We release the dataset, adapter, training notebooks, and the full evaluation harness so the entire pipeline---including both the reliability finding and the dataset audit that explains it---can be reproduced from a single 16~GB GPU. The broader lesson is that distribution-similarity metrics can move opposite to ground-truth reliability, and that fine-tuning small models on unverified synthetic data demands evaluation that does not depend on the synthetic distribution it was trained to match.

\section*{Declarations}

\textbf{Funding} \\
The authors declare that no funds, grants, or other support were received during the preparation of this manuscript.

\textbf{Ethics, Consent to Participate, and Consent to Publish} \\
All training and evaluation data are synthetic, generated by the Gemini~1.0~Pro API from prompt templates written by the author. No human-subjects data, personally identifiable information, or real student--advisor interactions are included. Any future real-user study (Section~5.3) will be conducted under an approved IRB protocol and reported separately.

\textbf{Use of Generative AI} \\
Generative-AI systems were used in three distinct roles, disclosed here for transparency: (i) the Gemini~1.0~Pro API generated the synthetic training and evaluation conversations (the \emph{subject} of study); (ii) a Sonnet-family model served as the blind LLM-as-judge, both in the model comparison (Section~4.3.2) and in the training-data prevalence audit (Section~\ref{sec:dataaudit}), an automatic evaluation tool whose limitations are stated in Section~5.2; and (iii) general-purpose LLMs assisted with paraphrasing and reference formatting during writing. All claims, the factuality verifications, the code, and the experimental design are the author's own. The fine-tuned model is a derivative of Mistral-7B-Instruct-v0.3 and is released under the original Mistral Research License.

\textbf{Responsible Release and Intended Use} \\
The released dataset and adapter are research objects that \emph{document} a failure mode; by this paper's own measurement they are factually unreliable on policy-sensitive advice and are \textbf{not} fit for deployment as a student-advising system. To prevent decontextualized reuse, the Hugging Face dataset card and model card carry the factuality finding (including the 28--40\% sampled-error estimate), a research-only intended-use statement, and an explicit prohibition on use for actual student advising without prior factuality-gating or retrieval grounding. The model's research-only Mistral Research License is consistent with, but does not substitute for, this intended-use restriction; institutional deployment in particular is discouraged because an institutional imprimatur would amplify the harm of confidently wrong advice.

\textbf{Competing Interests} \\
The author has no relevant financial or non-financial interests to disclose.

\textbf{Data and Code Availability} \\
The dataset (2{,}676 conversations, train + held-out test) is at \url{https://huggingface.co/datasets/millat/StudyAbroadGPT-Dataset}. The LoRA adapter and merged model are at \url{https://huggingface.co/millat/StudyAbroadGPT-7B-LoRa-Kaggle}. The training pipeline (Kaggle P100/T4 notebooks, training reports) is at \url{https://github.com/codermillat/StudyAbroadGPT}. Dataset generation and quality-control scripts are at \url{https://github.com/codermillat/study-abroad-dataset}. The evaluation harness---the 50-prompt blinded base-vs-LoRA outputs, automatic metrics, the reference-metric scripts, the LLM-as-judge scripts (\texttt{llm\_judge\_eval.py}, \texttt{analyze\_judge\_scores.py}), the per-pair judge scores, the source-verified model factuality catalog, and the dataset audit artifacts (the causal-match catalog and the 40-sample prevalence audit underlying Section~\ref{sec:dataaudit})---is released with this manuscript and reproduces every number in Section~4.

\end{document}